# Output Space Search for Structured Prediction

Janardhan Rao Doppa                                     DOPPA@EECS.OREGONSTATE.EDU
Alan Fern                                               AFERN@EECS.OREGONSTATE.EDU
Prasad Tadepalli                                        TADEPALL@EECS.OREGONSTATE.EDU
School of Electrical Engineering and Computer Science, Oregon State University, Corvallis, OR 97331, USA

## Abstract

We consider a framework for structured prediction based on search in the space of complete structured outputs. Given a structured input, an output is produced by running a time-bounded search procedure guided by a learned cost function, and then returning the least cost output uncovered during the search. This framework can be instantiated for a wide range of search spaces and search procedures, and easily incorporates arbitrary structured-prediction loss functions. In this paper, we make two main technical contributions. First, we define the limited-discrepancy search space over structured outputs, which is able to leverage powerful classification learning algorithms to improve the search space quality. Second, we give a generic cost function learning approach, where the key idea is to learn a cost function that attempts to mimic the behavior of conducting searches guided by the true loss function. Our experiments on six benchmark domains demonstrate that using our framework with only a small amount of search is sufficient for significantly improving on state-of-the-art structured-prediction performance.

## 1. Introduction

Structured prediction involves learning a predictor that can produce complex structured outputs given complex structured inputs. As an example, consider the problem of image scene labeling, where the structured input is an image and the structured output is a semantic labeling of the image regions. We study a new search-based approach to structured prediction. The approach involves first defining a combinatorial search space over complete structured outputs that allows for traversal of the output space. Next, given a structured input, a state-based search strategy (e.g. best-first or greedy search), guided by a learned cost function, is used to explore the space of outputs for a specified time bound. The least cost output uncovered by the search is then returned as the prediction.

The effectiveness of our approach depends critically on: 1) The identification of an effective combination of search space and search strategy over structured outputs, and 2) Our ability to learn a cost function for effectively guiding the search for high quality outputs. The main contribution of our work is to provide generic solutions to these two issues. First, we describe the limited-discrepancy search space, as a generic search space over complete outputs that can be customized to a particular problem by leveraging the power of (non-structured) classification learning algorithms. Second, we give a generic cost function learning algorithm that can be instantiated for a wide class of "ranking-based search strategies." The key idea is to learn a cost function that allows for imitating the search behavior of the algorithm when guided by the true loss function. We also provide experimental results for our approach on a number of benchmark problems and show that even when using a relatively small amount of search, the performance is comparable or better than the state-of-the-art.

## 2. Comparison to Related Work

A typical approach to structured prediction is to learn a cost function $\mathcal{C}(\mathbf{x}, \mathbf{y})$ for scoring a potential structured output $\mathbf{y}$ given a structured input $\mathbf{x}$. Given such a cost function and a new input $\mathbf{x}$, the output computation involves solving the so-called "Argmin" problem which is to find the minimum cost output for a given input. For example, the cost function is often represented as a linear model over template features of both $\mathbf{x}$ and $\mathbf{y}$ (Lafferty et al., 2001; Taskar et al., 2003; Tsochantaridis et al., 2004). Unfortunately exactly solving the Argmin problem is often intractable and efficient solutions exist only in limited cases such as when the dependency structure among features forms a tree. In such cases, one might simplify the features to allow for tractable inference or use heuristic optimization methods, which can be detrimental to prediction accuracy. In contrast, a potential advantage of our search-based approach is that it is relatively insensitive to the dependency structure of the features used to define the cost function. That is, the search procedure only needs to be able to evaluate the cost function at specific input-output pairs. Thus, we are free to increase the complexity of the cost function without considering its impact on the inference complexity. Another potential benefit of our approach is that since the search is over complete outputs, our inference is inherently an anytime procedure, meaning that it can be stopped at any time and return the best output discovered so far.

One approach to addressing inference complexity is cascade training (Felzenszwalb & McAllester, 2007; Weiss & Taskar, 2010; Weiss et al., 2010), where efficient inference is achieved by performing multiple runs of inference from a coarse to fine level of abstraction. Such approaches have shown good success, however, they place some restrictions on the form of the cost functions to facilitate "cascading." Another potential drawback of cascades and most other approaches is that they either ignore the loss function of a problem (e.g. by assuming Hamming loss) or require that the loss function be decomposable in a way that supports "loss augmented inference". Our approach is sensitive to the loss function and makes minimal assumptions about it, requiring only that we have a blackbox that can evaluate it for any potential output.

An alternative framework is classifier-based structured prediction. These algorithms avoid directly solving the Argmin problem by assuming that structured outputs can be generated by making a series of discrete decisions. The approach then attempts to learn a *recurrent classifier* that given a input **x** is iteratively applied in order to generate the series of decisions for producing the target output **y**. Simple training methods (e.g. (Dietterich et al., 1995)) have shown good success and there are some positive theoretical guarantees (Syed & Schapire, 2010; Ross & Bagnell, 2010). However, recurrent classifiers can be prone to error propagation (Kääriäinen, 2006; Ross & Bagnell, 2010). Recent work, e.g. SEARN (Hal Daumé III et al., 2009), SMiLe (Ross & Bagnell, 2010), and DAGGER (Ross et al., 2011), attempts to address this issue using more sophisticated training techniques and have shown state-of-the-art structured-prediction results. However, all these approaches use classifiers to produce structured outputs through a single sequence of greedy decisions. Unfortunately, in many problems, some decisions are difficult to predict by a greedy classifier, but are crucial for good performance. In contrast, our approach leverages recurrent classifiers to define good quality search spaces over complete outputs, which allows decision making by comparing multiple complete outputs and choosing the best.

The most closely related framework to ours is the SampleRank framework (Wick et al., 2011), which learns a cost function for guiding a type of Monte-Carlo search in the space of complete outputs. While it shares with our work the idea of explicit search in the output space, there are some significant differences. The SampleRank framework is focused on Monte-Carlo search, while our approach can be instantiated for a wide range of search algorithms. This is important since it is well understood in the search literature that the most appropriate type of search changes from problem to problem. In addition, the SampleRank framework is highly dependent on a hand-designed "proposal distribution" for guiding the search or effectively defining the search space. Rather, we describe a generic approach for constructing search spaces that is shown to be effective across a variety of domains.

## 3. Problem Setup

A structured prediction problem specifies a space of structure inputs $\mathcal{X}$, a space of structured outputs $\mathcal{Y}$, and a non-negative *loss function* $L : \mathcal{X} \times \mathcal{Y} \times \mathcal{Y} \mapsto \Re^+$ such that $L(x, y', y)$ is the loss associated with labeling a particular input $x$ by output $y'$ when the true output is $y$. We are provided with a training set of input-output pairs drawn from an unknown target distribution and the goal is to return a function/predictor from structured inputs to outputs whose predicted outputs have low expected loss with respect to the distribution. Since our algorithms will be learning cost functions over input-output pairs we assume the availability of a *feature function* $\Phi : \mathcal{X} \times \mathcal{Y} \mapsto \Re^n$ that computes an $n$ dimensional feature vector for any pair.

**Output Space Search.** We consider a framework for structured prediction based on state-based search in the space of complete structured outputs. The states of the search space are pairs of inputs and outputs $(x, y)$, representing the possibility of predicting $y$ as the output for $x$. A search space over those states is specified by two functions: 1) An *initial state function* $I$ such that $I(x)$ returns an initial search state for any input $x$, and 2) A *successor function* $S$ such that for any search state $(x, y)$, $S((x, y))$ returns a set of successor states $\{(x, y_1), \ldots, (x, y_k)\}$, noting that each successor must involve the same input $x$ as the parent.

Given a cost function $C$, that returns a numeric cost for any input-output pair (i.e. search state), we compute outputs using a search procedure (e.g. greedy search or beam search) guided by the cost function. In particular, given a input $x$, the search procedure starts at the initial state $I(x)$ and traverses the space according to some search strategy that is typically sensitive to $C$. After a specified amount of time, the search halts and the best state $(x, y')$ according to $C$ that was traversed is returned with $y'$ being the predicted output.

The effectiveness of our search-based approach depends on the quality of the search space defined by $I$ and $S$, the search strategy, and the quality of $C$. In the following sections, we describe our contributions toward defining effective search spaces and learning cost functions.

## 4. Search Spaces Over Complete Outputs

In this section we describe two search spaces over structured outputs: 1) The Flipbit space, a simple baseline, and 2) The limited-discrepancy search (LDS) space, which is intended to improve on the baseline. We start by describing recurrent classifiers, which are used in the definition of both spaces.

### 4.1. Recurrent Classifiers

A recurrent classifier constructs structured outputs based on a series of discrete decisions. This is formalized for a given structured-prediction problem by defining an appropriate *primitive search space*. It is a 5-tuple $\langle I, A, s, f, T \rangle$, where $I$ is a function that maps a input $x$ to an initial search node, $A$ is a finite set of actions (or operators), $s$ is the successor function that maps any search node and action to a successor search node, $f$ is a feature function from search nodes to real-valued feature vectors, and $T$ is the terminal state predicate that maps search nodes to $\{1, 0\}$ indicating whether the node is a terminal or not. Each terminal node in the search space corresponds to a *complete structured output*, while non-terminal nodes correspond to *partial structured outputs*. Thus, the decision process for constructing an output corresponds to selecting a sequence of actions leading from the initial node to a terminal. A recurrent classifier is a function that maps nodes of the primitive search space to actions, where typically the mapping is in terms of a feature function $f(n)$ that returns a feature vector for any search node. Thus, given a recurrent classifier, we can produce a output for $x$ by starting at the initial node of the primitive space and following its decisions until reaching a terminal.

As an example, for sequence labeling problems, the initial state for a given input sequence $x$ is a node containing $x$ with no labeled elements. The actions correspond to the selection of individual labels, and the successor function adds the selected label in the next position. Terminal nodes correspond to fully labeled sequences and the feature function computes a feature vector based on the input and previously assigned labels.

The most basic approach to learning a recurrent classifier is via *exact imitation*. For this, we assume that for any training input-output pair $(x, y)$ we can efficiently find an action sequence, or *solution path*, for producing $y$ from $x$. The exact imitation training approach learns a classifier by creating a classification training example for each node $n$ on the solution path of a structured example with feature vector $f(n)$ and label equal to the action followed by the path at $n$. Our experiments will use recurrent classifiers trained via exact imitation, but more sophisticated methods such as SEARN could also be used.

### 4.2. Flipbit Search Space

The *Flipbit search space* is a simple baseline space over complete outputs that uses a given recurrent classifier $h$ for bootstrapping the search. Each search state is represented by a sequence of actions in the primitive space ending in a terminal node representing a complete output. The initial search state corresponds to the actions selected by the classifier, so that $I(x)$ is equal to $(x, h(x))$, where $h(x)$ is the output generated by the recurrent classifier. The search steps generated by the successor function can change the value of one action at any sequence position of the parent state. In a sequence labeling problem, this corresponds to initializing to the recurrent classifier output and then searching over flips of individual labels. The flip-bit space is often used by local search techniques (without the classifier initialization) and is similar to the "search space" underlying Gibbs Sampling.

### 4.3. Limited-Discrepancy Search Space (LDS)

Notice that the Flipbit space only uses the recurrent classifier when initializing the search. The motivation behind our LDS space is to more aggressively exploit the recurrent classifier in order to improve the search space quality. LDS was originally introduced in the context of problem solving using heuristic search (Harvey & Ginsberg, 1995). To put LDS in context, we will describe it in terms of using a classifier for structured prediction given a primitive search space. If the learned classifier is accurate, then the number of incorrect action selections will be relatively small. However, even a small number of errors can propagate and cause poor outputs. The key idea behind LDS is to realize that if the classifier response was corrected at the small number of critical errors, then a much better output would be produced. LDS conducts a (shallow) search in the space of possible corrections in the hope of finding a solution better than the original.

More formally, given a classifier $h$ and its selected action sequence of length $T$, a discrepancy is a pair $(i, a)$ where $i \in \{1, \ldots, T\}$ is the index of a decision step and $a \in A$ is an action, which generally is different from the choice of the classifier at step $i$. For any set of discrepancies $D$ we let $h[D]$ be a new classifier that selects actions identically to $h$, except that it returns action $a$ at decision step $i$ if $(i, a) \in D$. Thus, the discrepancies in $D$ can be viewed as overriding the preferred choice of $h$ at particular decisions steps, possibly correcting for errors, or introducing new errors. For a structured input $x$, we will let $h[D](x)$ denote the output returned by $h[D]$ for the search space conditioned on $x$. At one extreme, when $D$ is empty, $h[D](x)$ simply corresponds to the output produced by the greedy classifier. At the other extreme, when $D$ specifies an action at each step, $h[D](x)$ is not influenced by $h$ at all and is completely specified by the discrepancy set. In practice, when $h$ is reasonably accurate, we will be primarily interested in small discrepancy sets relative to the size of the decision sequence. In particular, if the error rate of the classifier on individual decisions is small, then the number of corrections needed to produce a correct output will be correspondingly small. The problem is that we do not know where the corrections should be made and thus LDS conducts a search over the discrepancy sets, usually from small to large sets.

**Search Space Definition.** Given a recurrent classifier $h$, we define the corresponding limited-discrepancy search space over complete outputs as follows. Each search state in the space is represented as $(x, D)$ where $x$ is a structured input and $D$ is a discrepancy set. We view a state $(x, D)$ as equivalent to the input-output state $(x, h[D](x))$. The initial state function $I$ simply returns $(x, \emptyset)$ which corresponds to the original output of the recurrent classifier. The successor function $S$ for a state $(x, D)$ returns the set of states of the form $(x, D')$, where $D'$ is the same as $D$, but with an additional discrepancy. In this way, a path through the LDS search space starts at the output generated by the recurrent classifier and traverses a sequence of outputs that differ from the original by some number of discrepancies. Given a reasonably accurate $h$, we expect that high-quality outputs will be generated at relatively shallow depths of this search space and hence will be generated quickly.

### 4.4. Search Space Quality

Recall that in our experiments we train recurrent classifiers via exact imitation, which is an extremely simple approach compared to more elaborate methods such as SEARN. We now show the desirable property that the "exact imitation accuracy" optimized by that approach is directly related to the "quality" of the LDS search space, where quality relates the expected amount of search needed to uncover the target output. More formally, given a input-output pair $(x, y)$ we define the *LDS target depth* for an example $(x, y)$ and classifier $h$ to be the minimum depth of a state in the LDS space corresponding to $y$. Given a distribution over input-output pairs we let $d(h)$ denote the expected LDS target depth of a classifier $h$. Intuitively, the depth of a state in a search space is highly related to the amount of search time required to uncover the node (exponentially related for exhaustive search, and at least linearly related for more greedy search). Thus, we will use $d(h)$ as a measure of the quality of the LDS space. We now relate $d(h)$ to the classifier error rate.

For simplicity, assume that all decision sequences for the structured-prediction problem have a fixed length $T$ and consider a input-output pair $(x, y)$, which has a corresponding sequence of actions that generate $y$. Given a classifier $h$, we define its *exact imitation error* on $(x, y)$ to be $e/T$ where $e$ is the number of mistakes $h$ makes at nodes along the action sequence of $(x, y)$. Further, given a distribution over input-output pairs, we let $\epsilon_{ei}(h)$ denote the expected exact imitation error with respect to examples drawn from the distribution. Note that the exact imitation training approach aims to learn a classifier that minimizes $\epsilon_{ei}(h)$. Also, let $\epsilon_r(h)$ denote the *expected recurrent error* of $h$, which is the expectation over randomly drawn $(x, y)$ of the Hamming distance between the action sequence produced by $h$ when applied to $x$ and the true action sequence for $(x, y)$. The error $\epsilon_r(h)$ is the actual measure of performance of $h$ when applied to structured prediction. Recall that due to error propagation it is possible that $\epsilon_r(h)$ can be much worse than $\epsilon_{ei}(h)$, by as much as a factor of $T$. Proposition 1 shows that $d(h)$ is related to $\epsilon_{ei}(h)$ rather than the potentially much larger $\epsilon_r(h)$.

**Proposition 1.** *For any classifier $h$ and distribution over structured input-outputs, $d(h) = T \epsilon_{ei}(h)$.*

*Proof.* For any example $(x, y)$ the depth of $y$ in $S_h$ is equal to the number of imitation errors made by $h$ on $(x, y)$. To see this, simply create a discrepancy set $D$ that contains a discrepancy at the position of each imitation error that corrects the error. This set is at a depth equal to the number of imitation errors and the classifier $h[D]$ will exactly produce the exact action sequence for producing $y$. The result follows by noting that the expected number of imitation errors is equal to $\epsilon_{ei}$. □

It is illustrative to compare this result with the Flipbit space. Let $d'(h)$ be the expected target depth in the Flipbit space of a randomly drawn $(x, y)$. It is easy to see that $d'(h) = T \epsilon_r(h)$ since each search step can only correct a single error and the expected number of errors of the action sequence at the initial node is $T \epsilon_r(h)$. Since in practice and in theory $\epsilon_r(h)$ can be substantially larger than $\epsilon_{ei}(h)$, this shows that the LDS space will often be superior to the baseline Flipbit space in terms of the expected target depth. Since this depth relates to the difficulty of search and learning, we can then expect the LDS space to be advantageous when $\epsilon_r(h)$ is larger than $\epsilon_{ei}(h)$. In our experiments, we will see that this is indeed the case.

## 5. Cost Function Learning

In this section, we describe a generic framework for cost function learning that is applicable for a wide range of search spaces and search strategies. This approach is motivated by our observation that for a variety of structured prediction problems, we can uncover a high quality output if we can guide the output-space search by the loss function with respect to the target output $y^*$. Since the target output is not available at testing time, we aim to learn a cost function that mimics the search behavior of the loss function on the training data. With an appropriate choice of hypothesis space of cost functions, good performance on the training data translates to good performance on the testing data.

We now precisely define the notion of "guiding the search" with a loss function. If the loss function can be invoked arbitrarily by the search procedure, then matching its performance would require the cost function to approximate it arbitrarily closely, which is needlessly complex in most cases. Hence, we restrict ourselves to ranking-based search defined as follows.

**Ranking-based Search.** Let $\mathcal{P}$ be an anytime search procedure that takes an input $x \in \mathcal{X}$, calls a cost function $C$ over the pairs $\mathcal{X} \times \mathcal{Y}$ some number of times and outputs a structured output $y_{best} \in \mathcal{Y}$. We say that $\mathcal{P}$ is a ranking-based search procedure if the results of calls to $C$ are only used to compare the relative values for different pairs $(x, y)$ and $(x, y')$ with a fixed tie breaker. Each such comparison with tie-breaking is called a ranking decision and is characterized by the tuple $(x, y, y', d)$, where $d$ is a binary decision that indicates $y$ is a better output than $y'$ for input $x$. When requested, it returns the best output $y_{best}$ encountered thus far as evaluated by the cost function.

Note that the above constraints prohibit the search procedure from being sensitive to the absolute values of the cost function for particular search states $(x, y)$ pairs, and only consider their relative values. Many typical search strategies such as greedy search, best-first search, and beam search satisfy this property.

A *run* of a ranking-based search is a sequence $x, s_1, o_1, \ldots, s_n, o_n, y$, where $x$ is the input to the predictor, $y$ is the output, and $s_i$ is the internal memory state of the predictor just before the $i^{th}$ call to the ranking function. $o_i$ is the $i^{th}$ ranking decision $(x_i, y_i, y_i', d_i)$.

Given a hypothesis space $\mathcal{H}$ of cost functions, the cost function learning works as follows. It runs the search procedure $\mathcal{P}$ on each training example $(x, y^*)$ for a maximum time of $T_{max}$ substituting the loss function $L(x, y, y^*)$ for the cost function $C(x, y)$. For each run, it records the set of all ranking decisions $(x_i, y_i, y_i', d_i)$. The set of all ranking decisions from all the runs is given as input to a binary classifier, which finds a cost function $C \in \mathcal{H}$, consistent with the set of all such ranking decisions. The ranking-based search can be viewed as a Markov Decision Process (MDP), where the internal states of the search procedure correspond to the states of the MDP, and the ranking decision is an action. The following theorem can be proved by adapting the proof of (Fern et al., 2006) with minor changes, e.g., no discounting, and two actions, and applies to stochastic as well as deterministic search procedures.

**Theorem 1.** *Let $\mathcal{H}$ be a finite class of ranking functions. For any target ranking function $h \in H$, and any set of $m = \frac{1}{\epsilon} \ln \frac{|H|}{\delta}$ independent runs of a rank-based search procedure $\mathcal{P}$ guided by $h$ drawn from a target distribution over inputs, there is a $1 - \delta$ probability that every $\hat{h} \in \mathcal{H}$ that is consistent with the runs satisfies $L(\hat{h}) \leq L(h) + 2\epsilon L_{max}$, where $L_{max}$ is the maximum possible loss of any output.*

Although the theoretical result assumes that the target cost function $h$ is in the hypothesis space, in practice this is not guaranteed. To minimize the chances of not being able to find a consistent hypothesis, we will only include a smaller set of ranking decisions that are sufficient to preserve the best output of the algorithm at any time step. Since these decisions are specific to every search procedure, we will describe our approach on two specific search algorithms: greedy search and best-first beam search.

**Greedy Search:** In greedy search, at each search step $i$, only the best open (unexpanded) node $y_i$ and the best output $y_i^*$ uncovered so far as evaluated by the loss function are remembered. At each level $i$, we include decisions that rank $y_i$ higher than all its siblings, and $y_i^*$ higher than $y_{i-1}^*$.

**Best-first Beam Search:** In best-first beam search, at any search step $i$, a set of $b$ open nodes $B_i$ and the best output $y_i^*$ encountered so far are maintained, where $b$ is the beam width. The best open node $y_i \in B_i$ is expanded, and $B_{i+1}$ is computed to be the best $b$ nodes after expansion. The relevant ranking decisions ensure that all outputs in $B_i$ are ranked higher than those in $C_i \setminus B_i$, $y_i$ is ranked higher than every output in $B_i \setminus y_i$ and $y_i^*$ is ranked higher than $y_{i-1}^*$.

To further reduce the number of constraints considered by the learner, we do the following for both greedy search and beam search. Ranking constraints for exact imitation were generated until reaching $y^*$, the correct output, and after that we only generate constraint(s) to rank $y^*$ higher than best cost open node(s) as evaluated by the current cost function and continue the search guided by the cost function.

## 6. Summary of Overall Approach

Our approach consists of two main components, a recurrent classifier and a cost function, and we train them sequentially. First, we train the recurrent classifier as described in Section 4.1. We then use this trained classifier to define one of the two search spaces over complete outputs $S$ (either Flipbit or LDS) for every training input $x$ (see Section 4). Second, we train the cost function to score outputs for a given combination of search space over complete outputs $S$ and a search procedure $P$ as described in Section 5.

At test time, we use the learned recurrent classifier and cost function to make predictions as follows. For each test input $x$, we define the search space over complete outputs $S$ using the recurrent classifier and execute the search procedure $P$ in this search space guided by the cost function for a specified time bound. We return the best cost output $y$ that is uncovered during the search as the prediction for $x$.

## 7. Experiments and Results

**Datasets.** We evaluate our approach on the following six structured prediction problems (five benchmark sequence labeling problems and a 2D image labeling problem): **1) Handwriting Recognition (HW).** The input is a sequence of binary-segmented handwritten letters and the output is the corresponding character sequence $[a-z]^+$. This dataset contains roughly 6600 examples divided into 10 folds (Taskar et al., 2003). We consider two different variants of this task as in (Hal Daumé III et al., 2009), in HW-Small version, we use one fold for training and remaining 9 folds for testing, and vice-versa in HW-Large. **2) NETtalk Stress.** The task is to assign one of the 5 stress labels to each letter of a word. There are 1000 training words and 1000 test words in the standard dataset. We use a sliding window of size 3 for observational features. **3) NETtalk Phoneme.** This is similar to NETtalk Stress except that the task is to assign one of the 51 phoneme labels to each letter of the word. **4) Chunking.** The goal in this task is to syntactically chunk English sentences into meaningful segments. We consider the full syntactic chunking task and use the dataset from the CONLL 2000 shared task[1], which consists of 8936 sentences of training data and 2012 sentences of testing data. **5) POS tagging.** We consider the tagging problem for English language, where the goal is to assign the part-of-speech tag for each word in the sentence. The standard data from Wall Street Journal (WSJ) corpus[2] was used in our experiments. **6) Scene labeling.** This dataset contains 700 images of outdoor scenes (Vogel & Schiele, 2007). Each image is divided into patches by placing a regular grid of size 10×10 and each patch takes one of the 9 semantic labels (*sky, water, grass, trunks, foliage, field, rocks, flowers, sand*). Simple appearance features like color, texture and position are used to represent each patch. Training was performed with 600 images and the remaining 100 images were used for testing.

For all sequence labeling problems, the recurrent classifier labels a sequence using a left-to-right ordering and for scene labeling problem with an ordering from top-left to right-bottom in a row-wise raster form. To train the recurrent classifier, the output label of previous token is used as a feature to predict the label of the current token for all sequence labeling problems with the exception of chunking and POS tagging, where labels of two previous tokens were used. For scene labeling, the labels of neighborhood patches were used. In all our experiments, we train the recurrent classifier using exact imitation (see Section 4) via Perceptron for 100 iterations with learning rate 1. Prediction accuracy is measured with $F_1$ loss for the chunking task and Hamming loss for all the remaining tasks.

[1] http://www.cnts.ua.ac.be/conll2000/chunking/
[2] http://www.cis.upenn.edu/ treebank/

In all cases, the cost function over input-output pairs is second order, meaning that it is has features over neighboring label pairs and triples along with features of the structured input. We trained the cost function, as described in Section 5, in an online manner via Perceptron updates with learning rate 0.01 for 500 iterations (i.e., ranking constraints were generated on-the-fly in every iteration).

**Learners.** We report results for several instantiations of our framework. First, we consider our framework using a greedy search procedure for both the LDS and flip-bit spaces, denoted by **LDS-Greedy** and **FB-Greedy**. In both training and testing, the greedy search was run for a number of steps equal to the length of the sequence. Using longer runs did not impact results significantly. Second, we performed best-first beam search with a beam width of 100 in both the LDS and flib-bit spaces, denoted **LDS-BST-b100** and **FB-BST-b100**. The best-first search was run for 200 expansions in each case. We tried larger beam widths and search steps but performance was similar. Third, to see the impact of adding additional search at test time to a greedily trained cost function, we also used the cost function learned by LDS-Greedy and FB-Greedy in the context of a best-first beam search (beam width = 100) at test time in both the LDS and flip-bit space, denoted by **LDS-BST(greedy)** and **FB-BST(greedy)**. We also report the performance of recurrent classifier (**Recurrent**) and the exact imitation accuracy $(1 - \epsilon_{ei})$, which as described earlier are related to the structures of the flip-bit and LDS spaces.

We compare our results with other structured prediction algorithms including **CRFs** (Lafferty et al., 2001), **SVM-Struct** (Tsochantaridis et al., 2004), **SEARN** (Hal Daumé III et al., 2009) and **CASCADES** (Weiss & Taskar, 2010). For these algorithms, we report the best published results whenever available. In the remaining cases, we used publicly available code or our own implementation to generate those results. Ten percent of the training data was used to tune hyper-parameters. CRFs were trained using SGD[3]. SVM$^{hmm}$ was used to train SVM$_{struct}$ and the value of parameter $C$ was chosen from $\{10^{-4}, 10^{-3}, \cdots, 10^3, 10^4\}$ based on the validation set. Cascades were trained using the implementation[4] provided by the authors, which can be used for sequence labeling problems with Hamming loss. For SEARN we report the best published results with a linear classifier (i.e., linear SVMs instead of Perceptron) as indicated by $^B$ in the table and otherwise ran our own implementation of SEARN with optimal approximation as described in (Hal Daumé III et al., 2009) and optimized the interpolation parameter $\beta$ over the validation set. Note that we do not compare our results to SampleRank due to the fact that its performance is highly dependent on the hand-designed proposal distribution, which varies from one domain to another.

[3] http://leon.bottou.org/projects/sgd
[4] http://code.google.com/p/structured-cascades/

Table 1. Prediction accuracy results of different structured prediction algorithms.

| ALGORITHMS | DATASETS | | | | | | |
|---|---|---|---|---|---|---|---|
| | HW-Small | HW-Large | Stress | Phoneme | Chunk | POS | Scene labeling |
| $1 - \epsilon_{ei}$ | 73.9 | 83.99 | 77.97 | 77.09 | 88.84 | 92.5 | **78.61** |
| Recurrent | 65.67 | 74.87 | 72.82 | 73.58 | 88.51 | 92.15 | 56.64 |
| LDS-Greedy | 83.93 | 92.94 | 79.12 | 80.9 | 94.73 | **96.95** | 74.75 |
| FB-Greedy | 81.83 | 90.76 | 78.8 | 79.79 | 93.97 | 96.89 | 68.93 |
| LDS-BST(greedy) | **84.14** | **93.23** | 79.35 | 81.04 | 94.74 | 96.95 | 76.91 |
| FB-BST(greedy) | 81.83 | 90.76 | 78.8 | 79.83 | 94.05 | 96.89 | 69.25 |
| LDS-BST-b100 | 83.28 | 92.83 | **79.81** | **81.57** | 94.6 | 96.8 | 76.63 |
| FB-BST-b100 | 81.57 | 90.13 | 79.27 | 80.29 | 93.84 | 96.74 | 69.11 |
| CRF | 80.03 | 86.89 | 78.52 | 78.91 | **94.77** | 96.84 | - |
| SVM-Struct | 80.36 | 87.51 | 77.99 | 78.3 | 93.64 | 96.81 | - |
| SEARN | $82.12^B$ | $90.58^B$ | 76.15 | 77.26 | $94.44^B$ | 95.83 | 62.31 |
| CASCADES | 69.62 | 87.95 | 77.18 | 69.77 | - | 96.82 | - |

**Comparison to State-of-the-Art.** Table 1 shows the prediction accuracies of the different algorithms ('-' indicates that we were not able to generate results for those cases). Across all benchmarks we see that even the most basic instantiations of our framework, LDS-Greedy and FB-Greedy, produce results that are comparable or significantly better than the state-of-the-art. This is particularly interesting, since these results are achieved using a relatively small amount of search and the simplest search method and results tend to be the same or better for our other instantiations. A likely reason that we are outperforming CRFs and SVM-Struct is that we use second-order features, while those approaches use first-order features, since exact inference with higher order features is too costly, especially during training. As stated earlier, one of the advantages of our approach is that we can use higher-order features with negligible overhead.

To see whether our approach can benefit from further increasing the feature order, we generated results for our approach and Cascades using third-order features (not shown in table) for the NET-Talk and handwriting domains. Cascades improved over the results with second-order cost function for the handwriting dataset (81.87 for HW-Small and 93.76 for HW-Large), but degraded for the NET-Talk datasets. Our results with a third order cost function improved in both cases and are better than Cascades for the handwriting task (86.59 for HW-Small and 95.04 for HW-Large).

Finally, the improvement in the scene labeling domain is the most significant, where SEARN achieves an accuracy of 62.31 versus 74.75 for LDS-Greedy. In this domain, most prior work has considered the simpler task of classifying entire images into one of a set of discrete classes, but to the best of our knowledge no one has considered a structured prediction approach for patch classification. The only reported result for patch classification that we are aware of (Vogel & Schiele, 2007) obtain an accuracy of 71.7 (versus our best performance of 76.91) with non-linear SVMs trained i.i.d. on patches using more sophisticated features than ours.

**Adding More Search.** We see that LDS-BST(greedy) and FB-BST(greedy) are generally the same or better than LDS-Greedy and FB-Greedy, with the biggest improvement in the challenging scene labeling domain, improving from 74.75 to 76.91. This shows that it can be an effective strategy to train using greedy search and then insert that cost function into a more elaborate search at test time for further improvement. We see similar results for LDS-BST-b100 and FB-BST-b100 where the cost function was trained using best-first beam search. There was significant improvement for the NET-Talk datasets and scene labeling compared to LDS-BST and FB-BST. This illustrates that the approach can effectively train using the more complex search strategy of best-first beam search. It is interesting to note that LDS-BST(greedy) and LDS-BST-b100 perform similarly. Both methods use the same best-first search procedure at test time, but differ in that one trains with greedy search and the other with best-first search. This shows that based on these results there is not a clear advantage to train-ing in the context of best-first search, though there are benefits at test time. This is a point that deserves further investigation in future work.

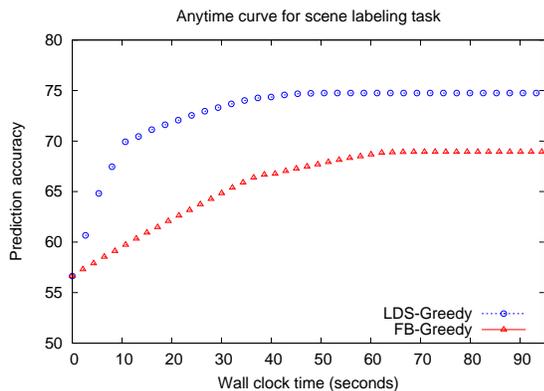

*Figure 1.* Anytime curves for scene labeling task comparing LDS-Greedy and FB-Greedy.

**LDS space vs. Flipbit space.** We see that generally the instances of our method that use the LDS space outperforms the corresponding instances that use the Flipbit space. Interestingly, if there is a large difference between the exact imitation accuracy $1 - \epsilon_{ei}$ and the recurrent classifier accuracy (e.g., Handwriting and Scene labeling), then the LDS space is significantly better than the flip-bit space. This is particularly true in our most complex problem of scene labeling where this difference is quite large, as is the gap between LDS and Flipbit.

Further, we compared the anytime curves between LDS-Greedy and FB-Greedy, which show the accuracy achieved by a method versus an inference time bound at prediction time. Generally we found that LDS-Greedy was comparable or better than the FB-Greedy curve and especially so for the Handwriting and Scene Labeling problems. Figure 1 shows the anytime curves for the Scene Labeling problem. We see that LDS-Greedy is dominant and improves accuracy much more quickly than FB-Greedy. For example, a 10 second time bound for LDS-Greedy achieves the same accuracy as FB-Greedy using 90 seconds. These results show the benefit of using the LDS space and empirically confirm our observations in Section 4 that the quality of the LDS and Flipbit spaces are related to the exact imitation and recurrent errors respectively.

## 8. Summary and Future Work

We studied a general framework for structured prediction based on search in the space of complete outputs. We showed how powerful classifiers can be leveraged to define an effective search space over complete outputs, and gave a generic cost function learning approach to score the outputs for any given combination of search space and search strategy. Our experimental results showed that a very small amount of search is needed to improve upon the state-of-the-art performance, validating the effectiveness of our framework. Future work includes studying robust training approaches to mitigate error propagation when the cost function is non-realizable and addressing scalability issues.

### Acknowledgements

This work was supported by NSF grants IIS 0964705, IIS 0958482, IIS-1018490 and DARPA contract FA8750-09-C-0179.